\documentclass[]{interact}

\usepackage{epstopdf}
\usepackage[caption=false]{subfig}

\usepackage[numbers,sort&compress]{natbib}
\bibpunct[, ]{[}{]}{,}{n}{,}{,}

\theoremstyle{plain}

\theoremstyle{definition}

\theoremstyle{remark}

\usepackage{todonotes}
\usepackage{diagbox}
\usepackage{longtable}
\usepackage{booktabs}
\usepackage{hyperref}
\usepackage[capitalise]{cleveref}
\usepackage{graphicx}       
\graphicspath{{images/}} 
\usepackage{tikz} 
\usepackage{subcaption}
\usepackage[justification=centering]{caption}
\usepackage{threeparttable} 
\usepackage{booktabs} 
\usepackage{float}
\usepackage{makecell}
\usepackage{soul}
\usepackage{float}
\usepackage{xspace}
\usepackage{adjustbox}
\usepackage{lineno}

\makeatletter
\DeclareRobustCommand\onedot{\futurelet\@let@token\@onedot}
\def\@onedot{\ifx\@let@token.\else.\null\fi\xspace}

\def\eg{\emph{e.g}\onedot} 
\def\ie{\emph{i.e}\onedot}

\def\etal{\emph{et al}\onedot}
\makeatother

\begin{document}


\title{Efficient Brood Cell Detection in Layer Trap Nests for Bees and Wasps: Balancing Labeling Effort and Species Coverage}

\author{
\name{Chenchang Liu\textsuperscript{1},
Felix Fornoff\textsuperscript{2},
Annika Grasreiner\textsuperscript{2},
Patrick Mäder\textsuperscript{1,3,5},
Henri Greil\textsuperscript{4},
and Marco Seeland\textsuperscript{1}\thanks{CONTACT Marco Seeland. Email: marco.seeland@tu-ilmenau.de}}
\affil{\textsuperscript{1}Technische Universität Ilmenau, Ehrenbergstraße 29, Ilmenau, Thüringen 98693, Germany;
\textsuperscript{2}Albert Ludwig University Freiburg, Stefan-Meier-Straße 76, Freiburg im Breisgau, Baden-Württemberg 79104, Germany;
\textsuperscript{3}Friedrich-Schiller-Universität Jena, Erbertstraße 1, Jena, Thüringen 07745, Germany;
\textsuperscript{4}Julius Kühn-Institut (JKI), Institute for Bee Protection, Federal Research Center for Cultivated Plants, Messeweg 11 \& 12, Braunschweig, Niedersachsen 38104, Germany;
\textsuperscript{5}German Centre for Integrative Biodiversity Research (iDiv), Halle-Jena-Leipzig, Leipzig, Sachsen 04103, Germany}
}

\maketitle

\begin{abstract}
Monitoring cavity-nesting wild bees and wasps is vital for biodiversity research and conservation. Layer trap nests (LTNs) are emerging as a valuable tool to study the abundance and species richness of these insects, offering insights into their nesting activities and ecological needs. However, manually evaluating LTNs to detect and classify brood cells is labor-intensive and time-consuming. 

To address this, we propose a deep learning based approach for efficient brood cell detection and classification in LTNs. LTNs present additional challenges due to densely packed brood cells, leading to a high labeling effort per image. Moreover, we observe a significant imbalance in class distribution, with common species having notably more occurrences than rare species. Comprehensive labeling of common species is time-consuming and exacerbates data imbalance, while partial labeling introduces data incompleteness which degrades model performance.

To reduce labeling effort and mitigate the impact of unlabeled data, we introduce a novel Constrained False Positive Loss (CFPL) strategy. CFPL dynamically masks predictions from unlabeled data, preventing them from interfering with the classification loss during training. 

Experimental results demonstrate that our method improves detection performance, balances model accuracy and labeling effort, while also mitigating class imbalance.

\end{abstract}


\section{Introduction}\label{sec1}
\subsection{Layer Trap Nests (LTNs) in Insect Monitoring}

The decline in insect biomass~\cite{HallmannPONEDeclineInsc2017} and wild bee populations~\cite{ZattaraBeeDeclineOneEarth2021} have drawn considerable concerns about biodiversity and ecosystem health~\cite{WagnerBioDivSurvBSE2022}. 
Insects, such as bees and wasps, play a crucial role in maintaining ecosystem functions, including pollination and pest control. 
Understanding the factors driving their population declines and their responses to environmental changes necessitates effective monitoring of their populations. 
However, large-scale or long-term monitoring datasets remain limited, despite their significant scientific and societal importance.
For instance, the European Union Pollinator Monitoring Scheme (EU PoMS)~\cite{EUPoMSProposal2021} aims to establish a standardized, field-based monitoring system to provide robust data on the status and trends of pollinator populations across EU countries.

Trap nests for solitary cavity-nesting bees and wasps provide nesting opportunities for these species. 
The provisioned offspring in these nests can be counted and identified to calculate the abundance and species richness of these insects. 
This approach has been used for decades to study bee and wasp responses to all kinds of stressors and conditions~\cite{StaabTrapNestOverviewMEE2018} and is representative for the biodiversity of bees in the area~\cite{KlausBAEImpvBeeMoni2024, TscharntkeBioIndiTrapNestJAE1998}. 
Traditional trap nests are made from materials like bamboo or reed internodes, which are commonly opened and reared in the lab~\cite{TscharntkeBioIndiTrapNestJAE1998}. 
In contrast to traditional ``reed nests'', recent studies have increasingly utilized wooden boards covered with foil~\cite{FornoffDNABarcodingTrapNestICD2023, LindermannCitizenSciMoniBeeCSTP2024}. 
These boards are stacked in multiple layers, hence referred to as layer trap nests (LTNs). 
The layers allow non-destructive visual inspection of the nest content and photo documentation. 
They also facilitate the targeted extraction of nest materials, such as food resources or larvae, for DNA barcoding~\cite{FornoffDNABarcodingTrapNestICD2023} or the detection of pesticide residues~\cite{AlkassabEcotoxiComparBee2020} within the nests. 
The capability for non-invasive visual inspection and photo documentation has also led to the development of a comprehensive identification key~\cite{lindermannBienenNisthilfen2023}, integrated into ``ID-logics'', an application designed for students and citizen scientists. 
This application supports citizen-science-based monitoring projects~\cite{Fornoff2025pollenMNU}, enabling both species monitoring and the assessment of population growth at LTN installation sites, thereby promoting education and conservation efforts.

Visual identification and classification of the nests, brood cells, and other nest contents can be challenging, especially for non-experts~\cite{KalweitCitizenSciWildbienenBamberg2024}.
Neural networks offer a solution to such challenges and can be optimized to identify nest contents at specific taxonomic levels, depending on the distinctiveness of visual characteristics within a brood cell.
For example, \textit{Isodontia mexicana} is the only cavity-nesting grasshopper-hunting wasp in Europe and the sole species that uses grass as nesting material in trap nests. 
Its brood cells are highly distinctive within Europe, allowing identification at the species level.
In contrast, \textit{Eumenine} potter wasps are more diverse, with brood cells that often appear similar across species~\cite{LindermannCitizenSciMoniBeeCSTP2024}, making identification possible only at the subfamily level.

\begin{figure}[htb]
    \centering 
    \includegraphics[width = 380pt]{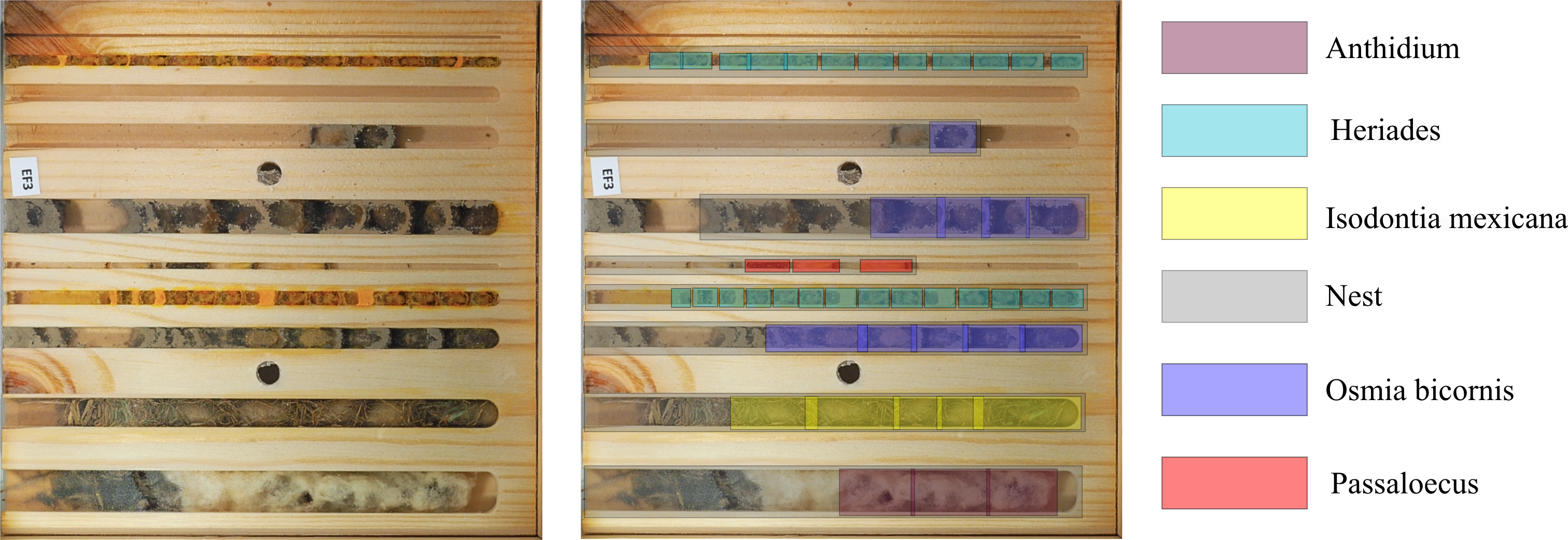}
    \centering
    \caption{\textbf{Example image of a layer trap nest (LTN).} Left: original image; right: image labeled with taxonomic information in terms of bounding boxes.}
    \label{fig:nesting_aid}
\end{figure}

Brood cells undergo significant changes in appearance throughout larval development. 
Initially, they are filled with food items, such as pollen or insects, which occupy most of the brood cell. 
As development progresses, larvae consume the food, transitioning into prepupae or forming cocoons for diapause. 
However, developmental errors can occur, halting progress and resulting in brood cells containing dead material. 
To achieve accurate classification, it is essential to account for the diverse species, species groups, and their various developmental stages, all of which may coexist within a single LTN. 
Due to differences in species abundance and the varying durations of developmental stages, the labeling process faces the challenge of balancing highly frequent classes with very rare ones within the training dataset.
In a typical LTN, as displayed in Fig.~\ref{fig:nesting_aid}, a variety of species are present. 
Among them, rare species with scarce occurrences are seldom observed and much harder to label comprehensively. 

The occurrence of species varies significantly and reflects the natural distribution of these species in the environment. 
Besides species occurrence frequencies, layer nests capture brood cells at all possible developmental stages of a cavity-nesting bee or wasp. 
Development can even be interrupted, which leads to a mix of developmental stages within one cavity.
For example, brood cells with pollen of which the larva died are visible next to brood cells with fully consumed pollen loads and a fully developed larva (\cref{fig:nesting_aid} cavity four from the bottom). 
Additionally, cavity nesting bees and wasps get attacked by a variety of natural enemies, including predators (\eg \textit{Trichodes}), parasitoids (\eg \textit{Ichneumonidae}), kleptoparasitic bees, known as cuckoo bees (\eg \textit{Stelis}), and kleptoparasitic wasps (\eg \textit{Chrysididae}).
Some of them are changing the appearance of a brood cell and can therefore be detected visually.

\subsection{Object Detection for LTNs and Labeling Effort}
Deep learning techniques are increasingly used for insect detection and analysis across various domains~\cite{ChenECE2019, MiaoSciRep2019}. 
Applying object detection algorithms to LTNs requires comprehensive labeling of brood cells~\cite{xu2019MissLabelsCVPR}. 
Unlabeled objects, treated as background during training, degrade model performance because the model learns to classify them as background. 
This requires labeling not only rare types of brood cells, referred to as \textit{minority} classes, but also common ones, referred to as \textit{majority} classes, that coexist in the same LTN image.
Thus, comprehensive labeling leads to redundant effort and class imbalance.

\subsection{Contributions}
We propose a deep learning based approach for brood cell detection and taxonomic classification in LTNs.  
To the best of our knowledge, this is the first work to apply deep learning for brood cell detection in LTNs.

LTN images show significant class imbalance, especially among common classes.  
To reduce labeling effort, we label up to 300 samples per class, leaving the remaining samples unlabeled.
To enhance detection performance in the presence of these unlabeled samples, we propose the \textbf{Constrained False Positive Loss (CFPL)} strategy.
CFPL dynamically masks predictions from unlabeled samples to selectively prevent them from interfering with the classification loss.
We evaluate the performance of CFPL on the collected LTN dataset and conduct an in-depth analysis of the results.


The remainder of this paper is structured as follows: \cref{sec:related-work} provides an overview of related work regarding LTNs and deep learning for ecological monitoring.
\cref{sec:methods} describes the methodology, including the dataset, detection loss function analysis, and our proposed CFPL method.
\cref{sec:experiments} presents the experimental setup, including data preparation, detection algorithm, and evaluation metrics.
\cref{sec:results} discusses the results of the experiments, and \cref{sec:conclusion} concludes the paper with a summary of the main findings and future research directions.

\section{Related Work}
\label{sec:related-work}

The monitoring of wild insects and their habitats using deep learning methods is an emerging field at the intersection of ecological research, computer vision, and artificial intelligence~\cite{BjergePONE2023, RoyPTRSBBS2024}. 
Research in this area not only demonstrates the potential of AI applications across different disciplines but also provides valuable data for ecological studies~\cite{BorowiecCDLEcoEvoDL2021}. 
This dual impact underscores the importance of advancing deep learning techniques for ecological monitoring, which has become increasingly popular in applications such as wildlife conservation, pollination support, agricultural sustainability, and insect control~\cite{BenahmedICE3ISDetectHoneybee2022, VermaTENCONInsectDetec2021, AlfarisyACMClsssiPests2018}.
In this section, we review existing works on deep learning for ecological monitoring, focusing on the use of layer trap nests as well as the detection and classification of wild bees and wasps in natural environments.

\subsection{Layer Trap Nests for Ecological Monitoring}
Trap nests for cavity-nesting bees and wasps have been used for decades to study their ecological dependencies, nesting preferences, and biotic interactions with natural enemies. 
More recently, wooden boards became popular in the commercial breeding of mason bees and investigations of mason bee responses to environmental stressors. 
In 2018, the first citizen science monitoring project for trap-nesting bees and wasps across Germany was launched~\cite{DuerrbaumMetabarcodingTrapNestME2023}.
Citizen science projects, such as~\cite{LindermannCitSciMonitorBees2024}, have used LTNs to monitor cavity-nesting bees and wasps in agricultural landscapes across Germany. 
The shift to layer trap nests has standardized visual identification, enabling automated image classification. 
Until today, LTN monitoring relies on scientists or at least trained citizen scientists to extract qualitative and quantitative information from images of LTN observations~\cite{LindermannCitSciMonitorBees2024, KalweitCitizenSciWildbienenBamberg2024}.

\subsection{Deep Learning for Insect Monitoring}
Deep learning has gained significant attention in insect classification and detection, particularly in the context of ecological monitoring, which involves the systematic collection and analysis of ecological data to track changes over time~\cite{spellerberg2005monitoring}.
Høye \etal~\cite{HyePNASEntomologyDL2021} demonstrated the use of deep learning for cost-effective insect monitoring through sensor-based methods and large-scale image classification. 
Spiesman \etal~\cite{SpiesmanSciRepDLBumbleBee} evaluated CNN models for bumble bee classification. 
They trained on over 89,000 images of 36 North American bumble bee species. 
Du \etal~\cite{DuALREntrMoniBee} developed a system capturing top and side-view photos of bees entering or exiting hives, enabling automated tracking, identification, and pollen analysis. 
Ratnayake \etal~\cite{RatnayakePOTrackingBee} proposed a hybrid detection-tracking algorithm for unmarked insects, combining background subtraction and deep learning to monitor insects even during occlusions. 
Bjerge \etal~\cite{BjergeRSECRealTimeInsect2021} introduced an intelligent camera system with real-time insect detection, tracking, and classification, providing insights into phenology, abundance, foraging, and movement ecology~\cite{CalvusEIInfieldMoniGroundNest2025}.

\subsection{Summary of Related Work}
Existing deep learning based approaches for insect monitoring have significantly advanced the detection of bees, wasps, and other insects in natural environments. 
However, most studies have primarily focused on monitoring free-roaming insects, which are mobile and may only be transient visitors to an area. 
In contrast, monitoring insect nests provides insights into species that are permanently present in a location, offering valuable information about the area's suitability for reproduction. 
To the best of our knowledge, automated methods for brood cell detection in LTNs remain unexplored, despite their potential to enhance ecological monitoring and conservation efforts.

\section{Methods}
\label{sec:methods}

\subsection{Dataset}

\subsubsection{Data Collection}
\label{sec:data_collection}
The LTN dataset originates from projects in Braunschweig, northern Germany, and Freiburg im Breisgau, southern Germany. 
LTNs from Braunschweig were installed in 2022 and 2023 along flower strips in the city of Braunschweig and the surrounding agricultural landscape, and the photos were captured in the winter of the same year using a Pentax digital camera with a macro lens.
LTNs in Freiburg consist of 176 layers made of spruce, beech, and Valchromat.
They were exposed in private gardens from April to October 2021, mounted on poles, and protected by plastic roofs.
These images were captured in winter 2021–2022 using an Olympus OM-D EM-1 camera with a macro lens. 
All LTNs feature square cavities of varying sizes (1–10 mm). 
In total, 712 images of unique LTNs are collected during one season of LTN monitoring.

\subsubsection{Data Labeling}
\label{sec:data_labeling}
We divide the images into three subsets: 70\% for training, 20\% for testing, and 10\% for validation.
Next, we manually labeled the LTN images using LabelStudio~\cite{LabelStudio}, a web-based open-source labeling tool.
The labeling process involves drawing bounding boxes around brood cells and assigning labels to each box.
The labels include taxonomic and status information.
Taxonomic information include species-level identification (\eg, \textit{Osmia bicornis}, \textit{Chelostoma florisomne}) or higher taxonomic groupings when species-level identification is not feasible (\eg, \textit{Heriades}, \textit{Trypoxylon}, \textit{Passaloecus}).  
Parasites like \textit{Cacoxenus indagator} and \textit{Trichodes apiarius} are also labeled. 
Status labels describe the brood cell's condition, such as ``Larva'', ``Hatched'', or ``Dead''.
\cref{tab:brood_cell_status} summarizes all status categories.
\begin{table}[htb]
\centering
\caption{Overview on brood cell status.}
\label{tab:brood_cell_status}
\begin{adjustbox}{width=1.0\linewidth}
\begin{tabular}{@{} c c l @{}} 
\toprule
\textbf{Status} & \textbf{Letter Code} & \textbf{Description} \\
\midrule
Dead     & D & Dead visible larva \\
Food     & F & Brood cell with only unconsumed food, no larva will develop \\
Hatched  & H & Brood cell with hatched bee or wasp traces (cocoon or exuvia) \\
Larva    & L & Visible alive larva \\
Prepupa  & P & Visible alive prepupa that stopped feeding \\
\bottomrule
\end{tabular}
\end{adjustbox}
\end{table}

\begin{figure}[htb]
    \centering 
    \includegraphics[width = 380pt]{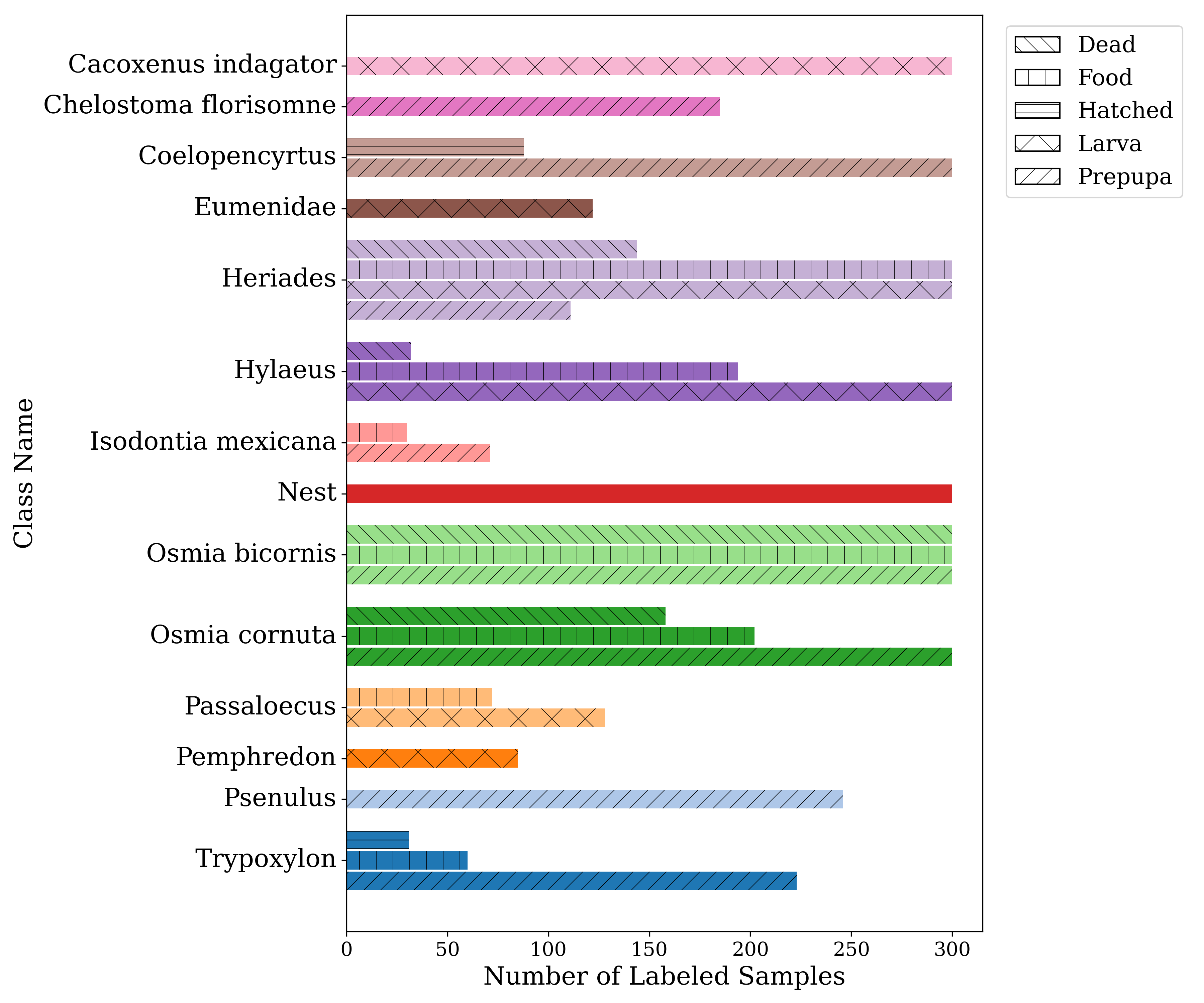}
    \centering
    \caption{Frequency of sample occurrences of the 28 most frequent taxa-status combinations.
    }
    \label{fig:class_distribution}
\end{figure}

The dataset demonstrates a significant imbalance in label distribution as shown in \cref{fig:class_distribution}.
While common species are very frequently observed and dominate the visual representation of images, rare species are seldom found and cover only a small proportion of an image, if at all.
To balance the labeling effort, we label up to 300 brood cell samples per class in the training set, even though some common classes contain far more than 300 samples.
Images from the testing and validation sets are fully labeled.
The training set is thereby divided into two groups of classes:
\begin{itemize}
    \item \textit{Majority}: they represent common species and have 300 labeled samples in the dataset. 
    More samples are available but remain unlabeled in the images.
    \item \textit{Minority}: they represent rare species and have fewer than 300 labeled samples in the dataset. 
    All samples are labeled.
\end{itemize}

The training set consists of 28 classes, with 10 \textit{majority} classes reaching beyond 300 samples each, and 18 classes \textit{minority} classes with fewer than 300 samples. 
In total, it includes 5'583 bounding boxes.  
On average, manually drawing a single bounding box in LabelStudio takes approximately 21 seconds, as measured during the labeling process. 
As a result, partially labeling the training set required approximately 33 hours. 
Based on the validation and test sets, each image is estimated to contain an average of 34 brood cells. 
Fully labeling all samples would have taken around 143 hours. 
Thus, the partial labeling approach saved an estimated 110 hours of human effort.

\subsection{Detection Algorithm}
The YOLO architecture is a fully convolutional neural network that divides the input image into a grid of cells, each of which predicts bounding boxes and class scores.
Unlike two-stage detectors, YOLO was the first one-stage detector and applies a single neural network to the entire image~\cite{RedmonCVPRYOLO2016}. 
This design allows YOLO to predict bounding boxes and class probabilities simultaneously, enabling real-time object detection without compromising accuracy~\cite{ZouObjDetecSurvey2023}. 

\subsection{YOLO Detection Loss}
The total loss in YOLO is a weighted sum of 2 main components: the classification loss, given by the Binary Cross-Entropy loss $\mathcal{L}_{\text{BCE}}$ for accurate object classification; and the regression loss, which includes Complete Intersection over Union loss $\mathcal{L}_{\text{CIoU}}$ and Distribution Focal Loss $\mathcal{L}_{\text{DLF}}$ for precise localization and sizing of bounding boxes~\cite{LiPICNIPSGeneralFocLoss2020}.
The loss function is computed as a weighted sum of these individual losses:
\begin{equation}
    \mathcal{L}_{\text{total}} = \lambda_{\text{CIoU}} \mathcal{L}_{\text{CIoU}} + \lambda_{\text{DFL}} \mathcal{L}_{\text{DFL}} + \lambda_{\text{BCE}} \mathcal{L}_{\text{BCE}},
    \label{eq:yolo_loss}
\end{equation}
where \(\lambda\) values are weights for balancing each loss component.

The classification loss, namely $\mathcal{L}_{\text{BCE}}$ is a performance measure for classification models that output predictions as probability values representing the likelihood of a data sample belonging to a specific class or category:
\begin{equation}
     \mathcal{L}_{\text{BCE}} = -\sum_{i=1}^{N} \sum_{c=1}^{N_c} \left[ y_{i,c} \log(p_{i,c}) + (1 - y_{i,c}) \log(1 - p_{i,c}) \right],
\end{equation}
where \(y_{i,c}\) is 1 if a sample belongs to class \(c\) or 0 if not.
\(p_{i,c}\) is the predicted probability of class \(c\), \(N\) is the number of labeled samples and \(N_c\) is the number of classes.

In the regression loss, $\mathcal{L}_\text{CIoU}$ measures the discrepancy between predicted bounding boxes and ground truth boxes. 
$\mathcal{L}_{\text{DFL}}$ estimates a probability distribution for bounding box locations, capturing uncertainties in object boundary predictions. 
This enhances the accuracy of bounding box localization, especially in scenarios where the boundaries of detected objects are ambiguous or difficult to determine.
    
YOLO divides the input image into a grid of cells, each of which predicts class scores and bounding boxes~\cite{RedmonCVPRYOLO2016, JiangPCSYOLORev2022}.
A class score is a logit representing the probability that a prediction belongs to a specific class.  
While the class scores contributes to the calculation of the classification loss $\mathcal{L}_{\text{BCE}}$, the bounding box predictions, \ie location, height, and width, contribute to the regression loss $\mathcal{L}_{\text{CIoU}}$ and $\mathcal{L}_{\text{DFL}}$.

In the regression loss computation, a ``foreground mask'' approach is employed to ensure that only ``positive'' regions, \ie, grid cells within ground truth objects and their related bounding box predictions, contribute to the regression loss components, $\mathcal{L}_{\text{CIoU}}$ and $\mathcal{L}_{\text{DFL}}$. 
The purpose of the ``foreground mask'' is to exclude predictions from backgrounds that do not contain objects. 
This approach ensures that only relevant bounding boxes, which contain objects of interest, contribute to the regression loss, improving the efficiency of object localization and boundary prediction~\cite{UltralyticsDocRef}. 

Conversely, when computing the classification loss, all predictions from all grid cells contribute. 
Predictions in ``positive regions'' learn to correctly classify objects.
Predictions in ``negative'' regions, \ie, regions not containing a labeled sample, are penalized as ``false positives'' and contribute to the classification loss as background.
However, when unlabeled samples exist in the ``negative regions'', such as brood cells from \textit{majority} classes that are not labeled, the classification loss mistakenly treats them as background.  
As a result, predictions on these unlabeled samples are incorrectly penalized as ``false positives'', leading to an increased classification loss and reduced model performance.  
This is particularly problematic for \textit{majority} classes, which contain a large number of such unlabeled common species instances.

\subsection{Constrained False Positive Loss (CFPL)}
\begin{figure}[htb]
    \centering 
    \includegraphics[width = 350pt]{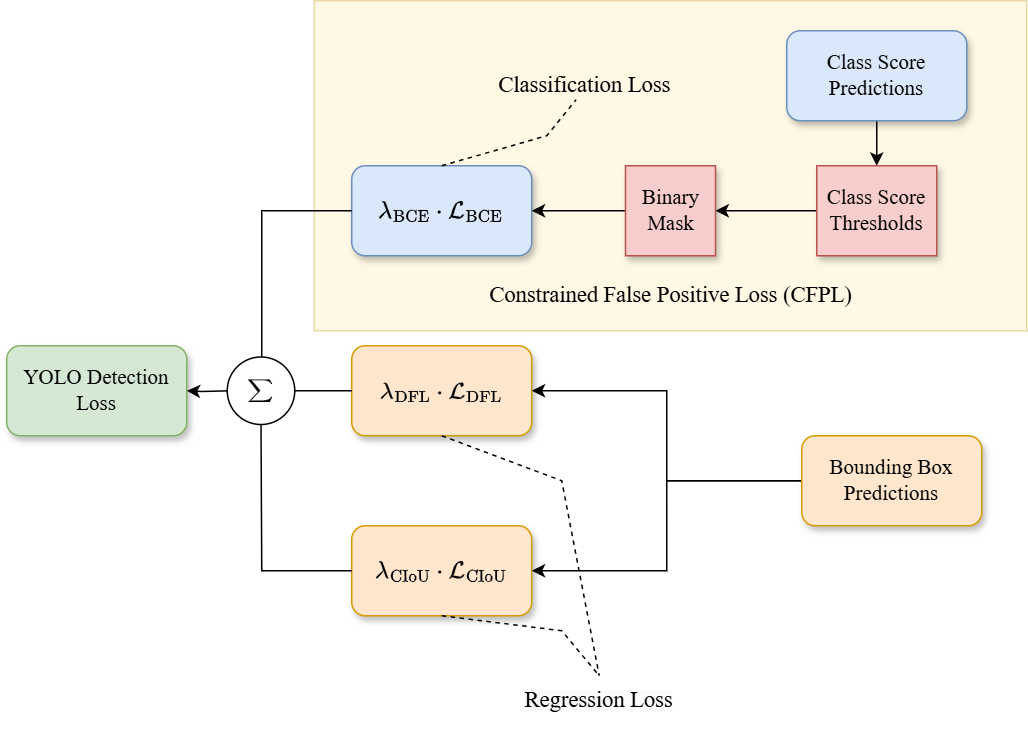}
    \caption{YOLO detection loss with CFPL.}
    \label{fig:loss_cfpl}
\end{figure}

We propose the Constrained False Positive Loss (CFPL) to mitigate the impact of unlabeled samples in YOLO's classification loss. 
The components of the YOLO detection loss in combination with CFPL are illustrated in \cref{fig:loss_cfpl}.

\begin{figure}[htb]
    \centering 
    \includegraphics[width = 390pt]{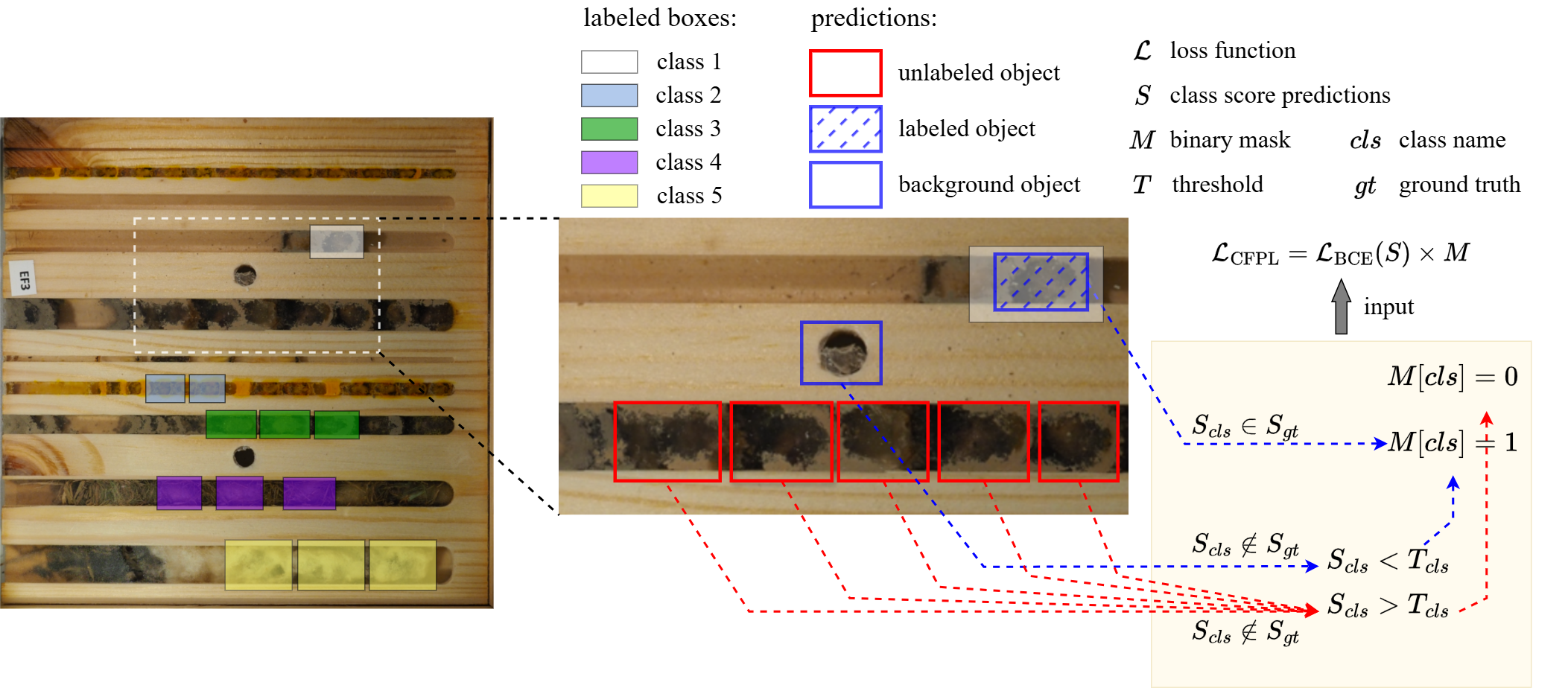}
    \caption{Constrained False Positive Loss (CFPL) with example image from the LTN dataset.}
    \label{fig:loss_cfpl_detailed}
\end{figure}

CFPL divides each image into three distinct areas and dynamically masks predictions from unlabeled areas:
\begin{enumerate}
    \item ground truth area: representing labeled bounding boxes;
    \item unlabeled area: containing unlabeled objects;
    \item background: areas without objects.
\end{enumerate}
\cref{fig:loss_cfpl_detailed} illustrates the implementation of CFPL in the YOLO detection loss.
When computing the classification loss, CFPL aims to exclude predictions from unlabeled areas, ensuring that only predictions from ground truth and background areas contribute to the loss calculation.
This is achieved by applying class-specific dynamic score thresholds to the classification loss, which are computed based on the predictions from the ground truth areas in each training iteration.
If a prediction exceeds the threshold, it is considered an unlabeled false positive and is masked out from the classification loss calculation.

The implementation of CFPL initializes a binary mask $\mathbf{M}$ that matches the dimensions of the class score predictions $\mathbf{S}$.
Class-specific predictions within ground truth areas $\mathbf{S}_{\text{cls}} \in \mathbf{S}_{\text{gt}}$ are flagged as valid in the mask, while predictions outside these areas $\mathbf{S}_{\text{cls}} \notin \mathbf{S}_{\text{gt}}$ are compared to the class-specific thresholds $T_{\text{cls}}$. 
Predictions exceeding the thresholds are flagged as invalid in the mask, where $\mathbf{M}[\text{cls}] = 0$, while those below it are retained as $\mathbf{M}[\text{cls}] = 1$. 
A whitelist of classes is used to select input classes, ensuring that CFPL is applied only to wanted classes, namely classes with unlabeled samples.
This way, CFPL's masking approach addresses the issue of interfering predictions caused by unlabeled samples, ensuring that only reliable predictions contribute to the classification loss, where $\mathcal{L}_\text{CFPL} = \mathcal{L}_\text{BCE}(\mathbf{S}) \times \mathbf{M}$.
The CFPL algorithm is provided as pseudocode in the Supplementary Material.

\section{Experiments}
\label{sec:experiments}

\subsection{Design}
The experiments aim to evaluate the feasibility of applying deep learning techniques for brood cell detection in LTN images and to assess the effectiveness of the proposed CFPL strategy in handling datasets with a significant number of unlabeled \textit{majority} samples. To achieve these goals, two experimental setups are conducted: one serving as a baseline using the original YOLOv8 for object detection, and the other incorporating YOLOv8 enhanced with CFPL for comparison. Each setup involves five independent training runs with different random seeds to ensure the robustness of the results.

\subsection{Data and Metrics}
The LTN dataset is prepared as described in \cref{sec:data_labeling}.
The distribution of classes is presented in the Supplementary Material.
Several metrics are employed to evaluate the performance of the model. 
\textit{Precision} quantifies the accuracy of the model’s positive predictions by measuring the proportion of true positives out of all positive predictions.
\textit{Recall} measures the proportion of true positive detections out of all actual positive samples, evaluating the model’s ability to detect all relevant objects.
For each object class, a \textit{precision-recall curve} is plotted by varying the \textit{Intersection over Union} threshold, and the \textit{Average Precision} (AP), defined as the area under the precision-recall curve, is computed. 

\subsection{Technical Information}
The YOLOv8 model from the Ultralytics library~\cite{JocherUltralytics2023} (version 8.3.28) is used throughout all experiments, implemented with Python 3.11.8 and PyTorch 2.5.1 and utilizing CUDA on an NVIDIA GeForce RTX 2080 Ti GPU. 
Each training session applies 120 epochs, with a batch size of 16 and an image size of 640 $\times$ 640. 
The AdamW optimizer is employed with an initial learning rate of 0.000313. 
In loss calculation, the weights are applied as default suggested by the Ultralytics library~\cite{JocherUltralytics2023}: the bounding box regression weight is set to 7.5, the distribution focal loss (DFL) weight to 1.5, and the classification loss weight to 0.5. 
Data augmentation techniques are applied to the training images, including random horizontal and vertical flipping, and brightness and contrast adjustments.

\subsection{Class-weighted Loss}
As discussed in \Cref{sec:data_labeling}, the dataset exhibits a significant class imbalance, particularly between the \textit{majority} and \textit{minority} classes.
To address this issue, we additionally tested using a \textit{class-weighted} classification loss $\mathcal{L}_{\text{BCE}}$ in Equation~\Cref{eq:yolo_loss}.
Class weights are calculated based on the inverse frequency of number of boxes in the training dataset for each class, giving higher weights to less frequent classes and lower weights to more frequent classes.
This approach can help to mitigate the impact of class imbalance by encouraging the model to pay more attention to less frequent classes during training.
However, we found that the application of a class-weighted loss did not lead to improvements in performance for either the \textit{majority} or \textit{minority} classes, and thus we did not include it in the final results presented in this paper.

\section{Results}
\label{sec:results}

\begin{figure}[H]
    \centering
    \includegraphics[width=0.9\textwidth]{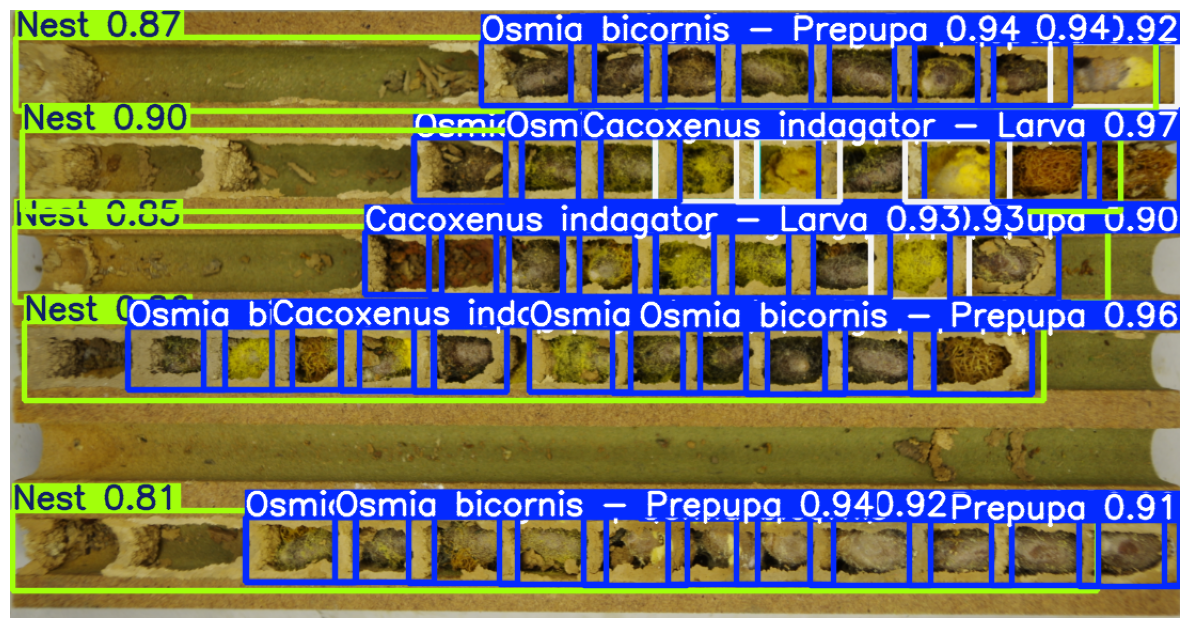}
    \caption{Exemplary detection results on an LTN.}
    \label{fig:example_detections}
\end{figure}

Exemplary detection results are shown in~\cref{fig:example_detections}. 
This example highlights the common species, \textit{Osmia bicornis} along with its respective statuses, \textit{Prepupa} and \textit{Larva}, both achieving confidence scores exceeding 0.9. 
The class \textit{Nest} is also identified in the example, which represents cavities in an LTN that are occupied by bees or wasps. 
All three classes belong to the \textit{majority} group.

We present the results for both \textit{majority} and \textit{minority} classes, focusing on Average Precision (AP) and recall at a confidence threshold of 0.5. 
A summary of the overall performance comparison between the two groups is provided in \cref{tab:overall_results}. 

\begin{table}[htb]
\centering
\caption{Aggregated performance metrics for \textit{majority} and \textit{minority} classes.}
\label{tab:overall_results}
\begin{adjustbox}{width=1.0\linewidth} 
\begin{tabular}{l|ccc|ccc}
\toprule
& \multicolumn{3}{c|}{Majority Classes} & \multicolumn{3}{c}{Minority Classes} \\
\cmidrule(lr){2-4} \cmidrule(lr){5-7}
Metric & Baseline & CFPL \textbf{(Ours)} & $\Delta_\text{abs}$ & Baseline & CFPL \textbf{(Ours)}& $\Delta_\text{abs}$ \\
\midrule
AP (\%)         & 63.11 $\pm$ 1.76 & \textbf{66.11} $\pm$ 3.00 & \textbf{3.00} & 44.29 $\pm$ 3.22 & \textbf{45.83} $\pm$ 2.11 & \textbf{1.54} \\
Recall (\%)     & 7.05 $\pm$ 2.16  & \textbf{46.54} $\pm$ 3.14 & \textbf{39.49} & 31.46 $\pm$ 3.64 & \textbf{35.57} $\pm$ 2.97 & \textbf{3.81} \\
\bottomrule
\end{tabular}
\end{adjustbox}
\end{table}

\begin{figure}[h!]
    \centering 
    \includegraphics[width = 390pt]{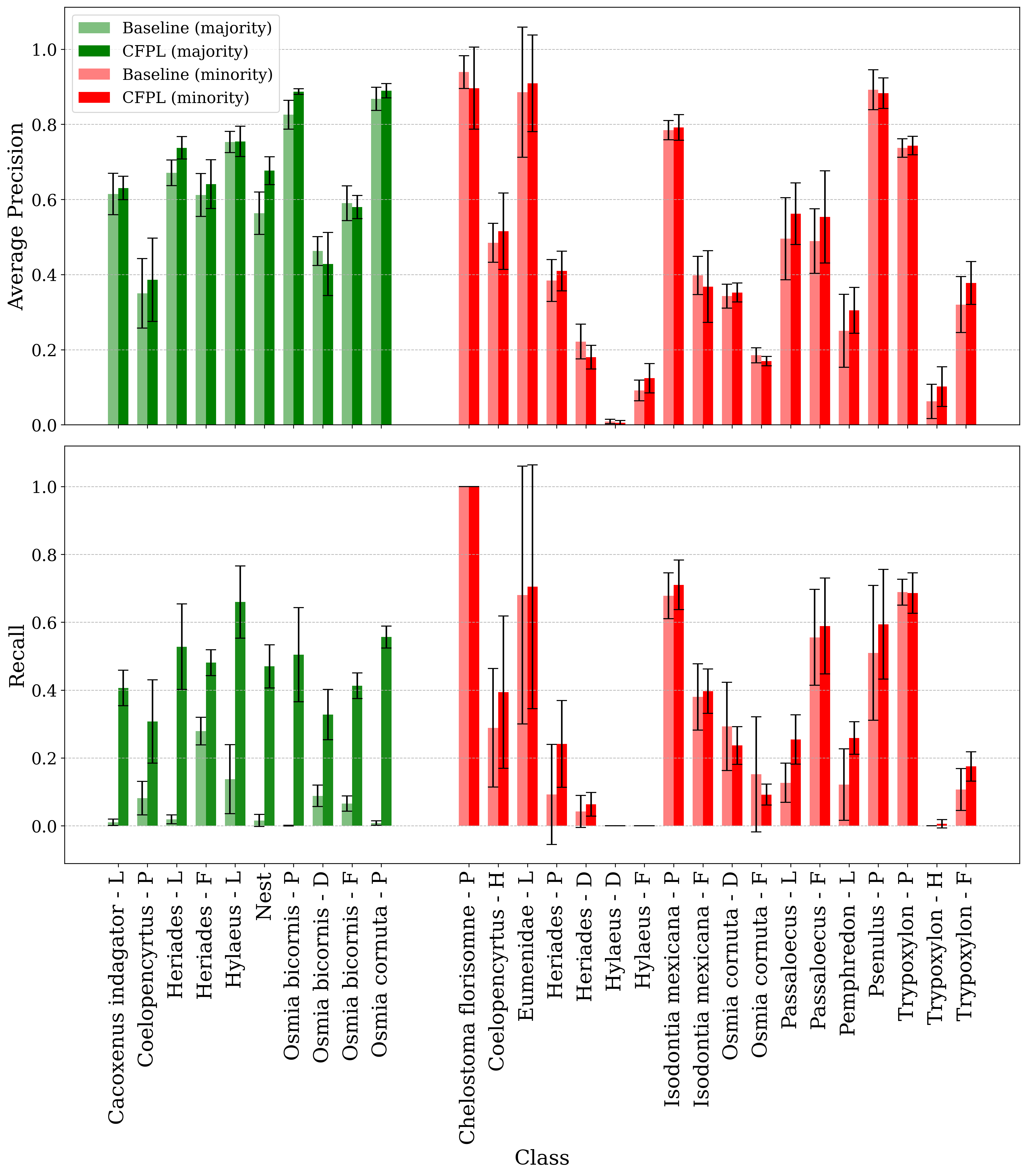}
    \caption{
       Comparison of AP and recall for all classes with and without CFPL.}
    \label{fig:combined_p_r_bar}
\end{figure}

The \textit{majority} classes achieved 63.11\% AP and 7.05\% recall in the baseline, improving to 66.11\% AP and 46.54\% recall with CFPL, a gain of 3.00\% in AP and 39.49\% in recall. 
For \textit{minority} classes, AP increased from 44.29\% to 45.83\%, and recall from 31.46\% to 35.57\% with CFPL.
The results indicate that automated brood cell detection in LTNs is feasible, and the proposed CFPL method effectively addresses the challenges posed by unlabeled samples in the dataset.
The results for the \textit{majority} and \textit{minority} classes are further analyzed in \cref{fig:combined_p_r_bar}, which presents the AP and recall values for each class, grouped by their respective categories.
More detailed statistical results for the individual classes within each group are available in the Supplementary Material.

In the baseline experiment, most \textit{majority} classes achieve an AP higher than 50\%, with an average AP of 63.11\% and the highest AP observed for the class \textit{Osmia cornuta - Prepupa} at 86.79\%. 
However, the recall in the baseline remains low, with only 7.05\% on average, and the highest of 13.74\% observed in the class \textit{Hylaeus - Larva}.
The low recall indicates that the model fails to detect a significant number of objects, which is a common issue in object detection tasks, especially when dealing with imbalanced datasets~\cite{ZouObjDetecSurvey2023}.
The model's performance is significantly improved with CFPL, as shown in \cref{fig:combined_p_r_bar}.
Compared to the baseline, an increase in both average precision (AP) and recall is observed for the \textit{majority} classes, with 3.00\% in AP and 39.49\% in recall, respectively. 
The most notable AP increase is observed for the \textit{Nest} class, with an increase of 11.32\%.
For the class \textit{Osmia cornuta - Prepupa}, the recall increases far more notably than the AP, with an increase of 54.85\%.
On average, the model is able to detect more than half of the samples, which is a significant improvement compared to the baseline.

With fewer than 300 training samples per class, the \textit{minority} classes represent relatively rare species. 
They exhibit mixed detection performance in the baseline, with a mAP of 44.29\% and a recall of 31.46\% (cf.~\cref{tab:overall_results}).
Some classes with more distinguishable features achieve both high AP and recall, such as \textit{Chelostoma florisomne - Prepupa}, \textit{Eumenidae - Larva}, and \textit{Psenulus - Prepupa}, each reaching over 80\% AP and over 60\% recall. 
Other classes, such as \textit{Hylaeus - Dead} and \textit{Trypoxylon - Hatched}, have both AP and recall values below 10\%.
These two classes have the fewest labeled samples, with only 31 and 32 instances, respectively.
CFPL is not applied to \textit{minority} classes.
Still, CFPL increases the average performance by 1.54\% in AP and 3.81\% in recall.
The performance increase can be attributed to CFPL’s strategy of filtering out false predictions from \textit{majority} classes, which shifts the model's focus towards the completely labeled \textit{minority} classes.

By analyzing the confusion matrix from the baseline experiments (cf. Supplementary Material), it is evident that the baseline results exhibit a high number of false negatives. 
Classes in the baseline are frequently misclassified as \textit{background}, particularly \textit{majority} classes with unlabeled samples. 
This aligns with the low recall observed in the baseline. 
For example, the misclassification rate of \textit{Cacoxenus indagator - Larva} as \textit{background} is 89\%, \textit{Nest} is 92\%, and \textit{Osmia bicornis - Prepupa} is 97\%. 
Similarly, false negatives are also present in the \textit{minority} classes, which have complete labeling, although the issue is less pronounced. 
For instance, \textit{Trypoxylon - Food} is misclassified as \textit{background} in 52\% of cases, and \textit{Isodontia Mexicana - Food} in 48\% of cases. 
Overall, the average misclassification rate as \textit{background} in the baseline is 56\%. 
These misclassifications occur because unlabeled data from the \textit{majority} classes confuses the model, leading it to incorrectly learn features of unlabeled objects as background. 
This results in a higher number of false negatives and ultimately lowers recall.

\begin{figure}[htb]
    \centering 
    \includegraphics[width = 390pt]{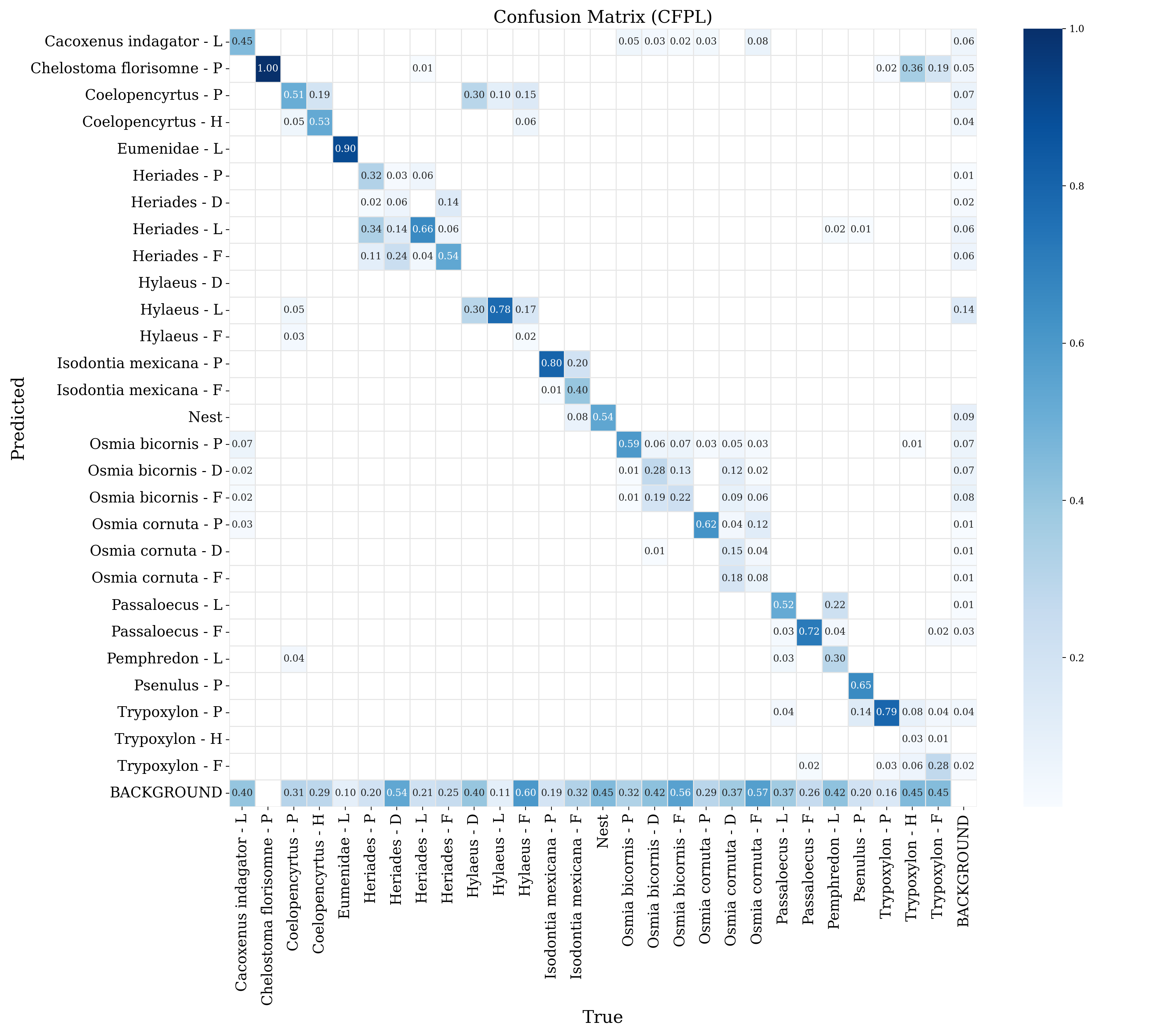}
    \caption{
        Confusion matrix with CFPL.
        }
    \label{fig:confusion_cfpl}
\end{figure}

The confusion matrix for the model trained with CFPL, shown in \cref{fig:confusion_cfpl}, reveals significant improvements compared to the baseline. 
The average misclassification rate as \textit{background} decreased from 0.56 to 0.33. 
For the \textit{majority} classes, misclassification rates as \textit{background} were notably reduced across all 10 classes. 
For example, the misclassification rate for \textit{Cacoxenus indagator - Larva} dropped from 0.89 in the baseline to 0.40 with CFPL. 
Similarly, the \textit{Nest} class experienced a reduction in its misclassification rate from 0.92 to 0.45. 
Improvements were also observed in the \textit{minority} classes. 
For instance, the misclassification rate for \textit{Heriades - Prepupa} as \textit{background} decreased from 0.53 in the baseline to 0.20 with CFPL, while for \textit{Trypoxylon - Food}, the rate dropped from 0.52 to 0.45. 
In summary, the implementation of CFPL significantly reduced the misclassification of objects as \textit{background}, leading to fewer false negatives and improved detection performance.

We use the confusion matrix as a reference for manual inspection based on expert knowledge and identify several main reasons for misclassification.
(1) In field environments, the content of a brood cell is often not fully visible.
This is the most frequent issue, as the insect may be enclosed within a cocoon, or be covered by paper, transparent film, or pollen.
Even when such exceptional cells are labeled correctly through manual effort, they exhibit highly diverse visual appearances, which increases confusion both among classes and with the background.
(2) It is evident that status identification is more challenging than taxon identification.
One representative example is that classes belonging to the \textit{Food} and \textit{Dead} groups are more frequently misclassified than others.
For instance, multiple errors occur for \textit{Osmia cornuta - Food} and \textit{Osmia cornuta - Dead}.
This is because both classes show visually ambiguous appearances at different stages of decay.
The \textit{Food} class represents unconsumed food, whose color, shape, and texture vary substantially over time, making it more difficult for the model to learn stable and consistent features.
Similarly, the \textit{Dead} class refers to dead larvae, whose appearance gradually change during degradation and resemble other classes.
(3) In addition, misclassifications are more likely to occur between species within the same genus.
Such confusion can be observed between \textit{Osmia cornuta} and \textit{Osmia bicornis}, or among species within the \textit{Heriades} genus.
These classes are visually highly similar and can be difficult to distinguish even with manual expert knowledge.

\section{Discussion}
Regarding the first objective, the results demonstrate that automated brood cell detection in LTNs is feasible using deep learning techniques.
The YOLO model achieves a mean Average Precision (mAP) of 66.11\% and a recall of 46.54\% for the \textit{majority} classes, indicating that the model can effectively detect brood cells in LTNs. 
For the \textit{minority} classes, the model achieves a mAP of 45.83\% and a recall of 35.57\%, indicating that the model can also detect rare species, albeit with notably lower performance.
The model's performance is further enhanced by the proposed CFPL method, which addresses the challenges posed by unlabeled samples in the dataset.
Next, we discuss the results in detail, focusing on the performance of \textit{majority} and \textit{minority} classes, and the impact of CFPL on the model's performance.

\subsection{Improved Detection with Reduced Labeling}
The introduction of CFPL leads to notable improvements in both AP and recall. 
For \textit{majority} classes, although only up to 300 samples per class are labeled, it is estimated that many more unlabeled samples exist in the dataset. 
Despite this, most \textit{majority} classes maintain high precision, as 300 labeled samples are sufficient for effective training. 
However, recall for \textit{majority} classes remains low in the baseline due to the presence of numerous unlabeled samples. 
Consequently, these classes are often misclassified as background, resulting in a higher number of false negatives and a significant drop in recall. 
CFPL effectively addresses this issue, improving recall across all \textit{majority} classes from an average of 7.05\% to 46.54\%. 
This demonstrates a more balanced trade-off between labeling effort and performance, primarily enhancing recall while also achieving a more consistent AP.

Two \textit{majority} classes, \textit{Osmia bicornis - Dead} and \textit{Osmia bicornis - Food}, exhibit slight decreases in AP, by 3.47\% and 1.02\%, respectively. 
An analysis of the confusion matrix reveals that the visual variability of these classes, combined with their interclass similarity, contributes to this decline. 
After implementing CFPL, predictions for these classes are more likely to be misclassified as other classes rather than as background. 
For instance, classes labeled as ``Food'' refer to food remnants, which can vary significantly in color, size, and state of decay, making them visually diverse. 
Similarly, the ``Dead'' class, representing deceased larvae, also exhibits substantial visual variability.

By excluding predictions for unlabeled samples of \textit{majority} classes during loss calculation, CFPL reduces the model's focus on these classes. 
This shift allows the model to allocate more attention to \textit{minority} classes, whose loss is fully considered. 
As a result, both recall and AP improve for \textit{minority} classes as well.

\begin{figure}[htb]
    \centering 
    \includegraphics[width = 370pt]{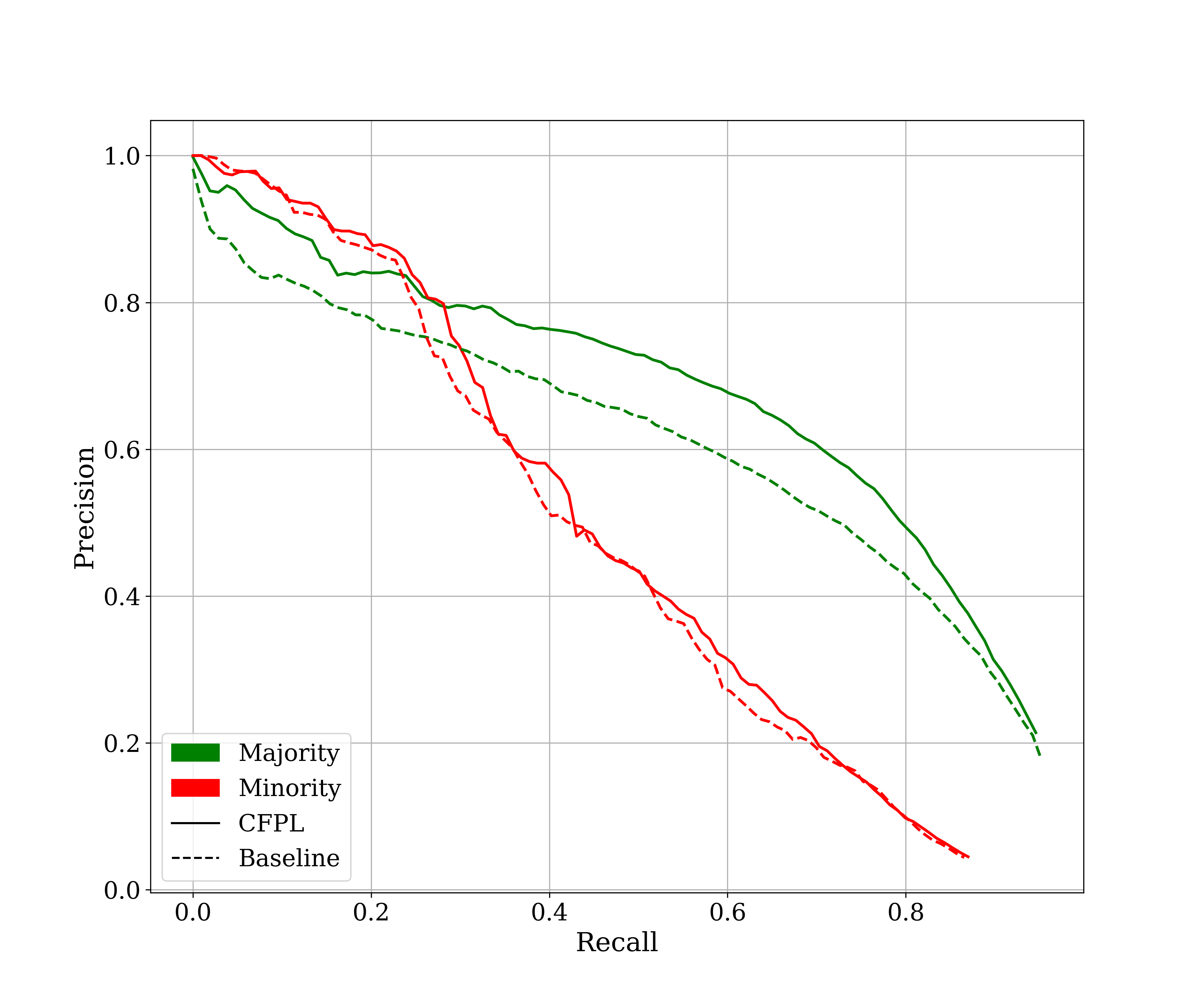}
    \caption{Precision-recall curves for each class group with and without CFPL.}
    \label{fig:p_r_curve}
\end{figure}

\Cref{fig:p_r_curve} compares the precision-recall curves for both class groups. 
Dashed lines represent the baseline model without CFPL, while solid lines depict the model with CFPL. 
The \textit{majority} classes are shown in green, and the \textit{minority} classes are shown in red. 
The CFPL approach effectively enhances performance by mitigating the impact of unlabeled data, resulting in a more balanced and consistent performance across both class groups.

In summary, CFPL consistently enhances model performance across both class groups. 
The significant increase in recall indicates a reduction in false negatives, which is particularly advantageous in real-world scenarios with a large number of unlabeled samples. 
This improvement demonstrates the model's ability to detect positive samples more effectively without confusing them with the background. 
The increase in AP reflects a more balanced overall performance, with gains in precision across all recall levels. 
Given the time-intensive nature of image labeling, especially for densely populated images (cf.~\cref{fig:example_detections}), these findings highlight that model performance can be substantially improved even with limited labeling effort.

The \textit{majority} classes maintain relatively higher precision even in the baseline. 
CFPL consistently helps to improve accurate detections across the recall range for \textit{majority} classes.
For \textit{minority} classes, the overall precision is notably lower compared to the \textit{majority} group.
Although the \textit{minority} classes do not contain any unlabeled samples, their performance still improve with CFPL.

\subsection{Limitations}
Despite the effectiveness of CFPL in reducing labeling effort, the limited dataset size and diversity represents a significant limitation of the current work.
The time and location of image collection, as discussed in \cref{sec:data_collection}, are key factors limiting the diversity of the dataset.
These constraints restrict both the overall dataset size and the model's ability to detect brood cells beyond the overwintering stage.
In particular, the limited number of training samples is a major factor contributing to the low performance observed for some \textit{minority} classes.
With only 712 images currently available, the dataset does not provide sufficient representation of rare taxa and their developmental status.
Incorporating a class-weighted loss strategy did not lead to improvements in performance for the \textit{minority} classes, suggesting that the underperformance is primarily limited by a lack of representative training samples rather than bias from class imbalance.
We therefore believe that increasing the dataset size, particularly for the \textit{minority} classes, is crucial for improving model performance in future work.
In addition to increasing the dataset size, strategies such as targeted data augmentation~\cite{LiuVISAPP2026} or focal loss~\cite{lin2017focal} could be explored in future work to further enhance performance for the \textit{minority} classes.

Another limitation arises from the selection of layer trap nests as the hardware setup for data collection.
While LTNs are widely used in ecological monitoring and provide a practical means of collecting data on bees and wasps, our dataset is limited to images captured from our specific hardware setup, which may not fully represent the diversity of real-world scenarios.
Our nesting aids are mostly made from homogeneous composite material, specifically high-density fiberboard (HDF), which is a highly dense wood-based material. 
However, trap nests made from naturally grown solid wood are also commonly used in real-world settings, yet such material is not represented in our dataset.
Thus, differences in material texture and color may confuse the model and limit its ability to generalize across different hardware setups.

Additionally, our images were captured under a relatively standardized setup, with fixed camera positions and viewing angles, as described in \cref{sec:data_collection}. 
More challenging cases, such as close-up images or images taken from diverse viewpoints and under varying lighting conditions, are not included in our dataset. 
This lack of diversity in the image data may limit the model's ability to generalize to different real-world scenarios, where images may be captured under a wider range of conditions.

We believe a more refined definition of brood cell status could further improve the analysis.
At present, remnants of brood cells are excluded, and only the five most distinctive status listed in \cref{tab:brood_cell_status} are defined. Furthermore, in real-world scenarios, even individuals of the same species may collect different building materials or pollen, which can result in substantial visual variation.

\section{Conclusion}
\label{sec:conclusion}

Trap-nesting bees and wasps serve as indicators of pollinator biodiversity~\cite{TscharntkeBioIndiTrapNestJAE1998}.
An automated detection model capable of identifying and classifying all species to the highest possible taxonomic resolution and accurately determining their developmental status would significantly contribute to the monitoring of this species group.
Building on this need, our work presents the first application of deep learning techniques to support the automated evaluation of layer trap nests (LTNs) for bees and wasps.

In addition, we proposed a novel Constrained False Positive Loss (CFPL) strategy to address the challenges posed by an imbalanced dataset and the significant labeling effort required for object detection tasks in densely populated images. 
CFPL dynamically masks predictions in unlabeled areas of an image, selectively preventing them from contributing to the classification loss. 
Our dataset is categorized into \textit{majority} and \textit{minority} classes, representing classes with over 300 and less than 300 occurrences.
Experiments are conducted on the same dataset with and without the implementation of CFPL.
Our results demonstrate that CFPL effectively enhances detection performance by mitigating the impact of unlabeled data, achieving an average recall improvement of 39.49\% and a 3.00\% improvement in average precision (AP) for incompletely labeled \textit{majority} classes.
Also the \textit{minority} classes benefit from CFPL, with an increase of 3.81\% in recall and a 1.54\% increase in AP.
These findings highlight a practical approach to data labeling, where high model performance can be achieved by labeling only a fixed subset of samples while reducing the need for exhaustive labeling of frequently occurring classes. 
This significantly reduces labeling effort while maintaining strong detection performance.

In conclusion, our work provides a novel perspective on reducing labeling efforts in deep learning based object detection, supporting large-scale ecological monitoring and conservation efforts with long tail class distributions. 
We demonstrate that a limited number of samples is sufficient for training, and our CFPL strategy can effectively improve performance with the existence of unlabeled samples.
Currently, the image data is constrained to the overwintering status of the brood cells and still lacks sufficient replications of rare species and classes. 
As part of future citizen-science-based monitoring research, images will be taken throughout the season, adding even further developmental stages or image classes to the database. 
Although the current model is limited by prediction accuracy and dataset size, it provides a solid foundation for further model development within ongoing monitoring projects.

\section*{Acknowledgments}
C.L., P.M., H.G., and M.S. were supported by the Federal Agency for Nature Conservation (BfN) through the project ``BeesUp'' (Grant Number: 3520685A29 and 3520685B29).
We would like to thank Felix Klaus, Jonah Krause, and Dennis Leer for setting up and evaluating the nesting aids. 
Our heartfelt thanks go to all the helpers who assisted in preparing the layer nests and taking photos: Jana Deierling, Jan Demuth, Jan Fritsch, Fredrik Mühlberger, Magdalena Podjaski, Niels Ketelsen, Simon Fischer, Peer Schneider, Dr. Abdulrahim Alkassab, Carmen Schröder, Nikola Kügler, Nikita Gubin, Michelle Kubiczek, Mattes Born, Hannah Hinsch, Lea Keudel, Xaver Habel, Hanna Gardein, Joris Husung. 
Special thanks go to Jonah Krause for labeling the brood cells in the photos.

\section*{Data Availability Statement}
The image dataset used in this work is available from Zenodo at \url{https://zenodo.org/records/20523520}.

\section*{Declaration of AI Usage}
The generative AI tool ChatGPT (versions 4.5, 5.4, and 5.5)~\cite{chatgpt2026} was used to support language editing and to assist with code during data preprocessing and result analysis. 
The authors manually checked all AI-generated content and take full responsibility for the correctness of the textual descriptions and the scientific content associated with this use.

\section*{Conflict of Interest}
The authors declare that they have no conflicts of interest.

\bibliographystyle{tfq}
\bibliography{interacttfqsample}

@INPROCEEDINGS{BenahmedICE3ISDetectHoneybee2022,
  author={Benahmed, Hadi Kouider and Bensaad, Mohamed Lahcen and Chaib, Noureddine},
  booktitle={2022 2nd International Conference on Electronic and Electrical Engineering and Intelligent System (ICE3IS)}, 
  title={Detection and tracking of honeybees using YOLO and StrongSORT}, 
  year={2022},
  volume={},
  number={},
  pages={18-23},
  doi={10.1109/ICE3IS56585.2022.10010142}}

@INPROCEEDINGS{VermaTENCONInsectDetec2021,
  author={Verma, Shani and Tripathi, Shrivishal and Singh, Anurag and Ojha, Muneendra and Saxena, Ravi R},
  booktitle={TENCON 2021 - 2021 IEEE Region 10 Conference (TENCON)}, 
  title={Insect Detection and Identification using YOLO Algorithms on Soybean Crop}, 
  year={2021},
  volume={},
  number={},
  pages={272-277},
  doi={10.1109/TENCON54134.2021.9707354}}

@inproceedings{AlfarisyACMClsssiPests2018,
author = {Alfarisy, Ahmad Arib and Chen, Quan and Guo, Minyi},
title = {Deep learning based classification for paddy pests \& diseases recognition},
year = {2018},
isbn = {9781450364201},
publisher = {Association for Computing Machinery},
address = {New York, NY, USA},
doi = {10.1145/3208788.3208795},
pages = {21–25},
numpages = {5},
location = {Chengdu, China},
series = {ICMAI '18}
}

@article{SpiesmanSciRepDLBumbleBee,
author = {Spiesman, Brian and Gratton, Claudio and Hatfield, Rich and Hsu, William and Jepsen, Sarina and McCornack, Brian and Patel, Krushi and Wang, Guanghui},
year = {2021},
month = {04},
pages = {},
title = {Assessing the potential for deep learning and computer vision to identify bumble bee species from images},
volume = {11},
journal = {Scientific Reports},
doi = {10.1038/s41598-021-87210-1}
}

@article{DuALREntrMoniBee,
author = {Du, Jingwen and Brothers, Zach and Valdes, Leah and Napp, N. and Petersen, Kirstin},
year = {2022},
month = {02},
pages = {},
title = {Automated entrance monitoring of managed bumble bees},
volume = {27},
journal = {Artificial Life and Robotics},
doi = {10.1007/s10015-022-00748-9}
}

@article{RatnayakePOTrackingBee,
author = {Ratnayake, Malika and Dyer, Adrian and Dorin, Alan},
year = {2021},
month = {02},
pages = {e0239504},
title = {Tracking individual honeybees among wildflower clusters with computer vision-facilitated pollinator monitoring},
volume = {16},
journal = {PLOS ONE},
doi = {10.1371/journal.pone.0239504}
}

@misc{ZouObjDetecSurvey2023,
      title={Object Detection in 20 Years: A Survey}, 
      author={Zhengxia Zou and Keyan Chen and Zhenwei Shi and Yuhong Guo and Jieping Ye},
      year={2023},
      eprint={1905.05055},
      archivePrefix={arXiv},
      primaryClass={cs.CV},
}

@article{JiangPCSYOLORev2022,
title = {A Review of Yolo Algorithm Developments},
journal = {Procedia Computer Science},
volume = {199},
pages = {1066-1073},
year = {2022},
issn = {1877-0509},
author = {Peiyuan Jiang and Daji Ergu and Fangyao Liu and Ying Cai and Bo Ma}
}

@misc{UltralyticsDocRef,
  title = {Reference for ultralytics/utils/loss.py},
  howpublished = {\url{https://docs.ultralytics.com/reference/utils/loss/}},
  note = {Accessed: 2024-07-08}
}

@inproceedings{LiPICNIPSGeneralFocLoss2020,
author = {Li, Xiang and Wang, Wenhai and Wu, Lijun and Chen, Shuo and Hu, Xiaolin and Li, Jun and Tang, Jinhui and Yang, Jian},
title = {Generalized focal loss: learning qualified and distributed bounding boxes for dense object detection},
year = {2020},
isbn = {9781713829546},
publisher = {Curran Associates Inc.},
address = {Red Hook, NY, USA},
booktitle = {Proceedings of the 34th International Conference on Neural Information Processing Systems},
articleno = {1763},
numpages = {11},
location = {Vancouver, BC, Canada},
series = {NIPS '20}
}

@article{LindermannCitSciMonitorBees2024,
author = {Lindermann, Lara and Grabener, Swantje and Hellwig, Niels and Stahl, Johanna and Dieker, Petra},
year = {2024},
month = {09},
pages = {22},
title = {Citizen Science-Based Monitoring of Cavity-Nesting Wild Bees and Wasps – Benefits for Volunteers, Insects, and Ecological Science},
volume = {9},
journal = {Citizen Science: Theory and Practice},
doi = {10.5334/cstp.632}
}

@article{HallmannPONEDeclineInsc2017,
  title = {More than 75 percent decline over 27 years in total flying insect biomass in protected areas},
  volume = {12},
  ISSN = {1932-6203},
  DOI = {10.1371/journal.pone.0185809},
  number = {10},
  journal = {PLOS ONE},
  publisher = {Public Library of Science (PLoS)},
  author = {Hallmann,  Caspar A. and Sorg,  Martin and Jongejans,  Eelke and Siepel,  Henk and Hofland,  Nick and Schwan,  Heinz and Stenmans,  Werner and M\"{u}ller,  Andreas and Sumser,  Hubert and H\"{o}rren,  Thomas and Goulson,  Dave and de Kroon,  Hans},
  editor = {Lamb,  Eric Gordon},
  year = {2017},
  month = oct,
  pages = {e0185809}
}

@article{BjergePONE2023,
  title = {Accurate detection and identification of insects from camera trap images with deep learning},
  volume = {2},
  ISSN = {2767-3197},
  DOI = {10.1371/journal.pstr.0000051},
  number = {3},
  journal = {PLOS Sustainability and Transformation},
  publisher = {Public Library of Science (PLoS)},
  author = {Bjerge,  Kim and Alison,  Jamie and Dyrmann,  Mads and Frigaard,  Carsten Eie and Mann,  Hjalte M. R. and Høye,  Toke Thomas},
  editor = {Fang,  Wei-Ta},
  year = {2023},
  month = mar,
  pages = {e0000051}
}

@article{RoyPTRSBBS2024,
  title = {Towards a standardized framework for AI-assisted,  image-based monitoring of nocturnal insects},
  volume = {379},
  ISSN = {1471-2970},
  DOI = {10.1098/rstb.2023.0108},
  number = {1904},
  journal = {Philosophical Transactions of the Royal Society B: Biological Sciences},
  publisher = {The Royal Society},
  author = {Roy,  D. B. and Alison,  J. and August,  T. A. and Bélisle,  M. and Bjerge,  K. and Bowden,  J. J. and Bunsen,  M. J. and Cunha,  F. and Geissmann,  Q. and Goldmann,  K. and Gomez-Segura,  A. and Jain,  A. and Huijbers,  C. and Larrivée,  M. and Lawson,  J. L. and Mann,  H. M. and Mazerolle,  M. J. and McFarland,  K. P. and Pasi,  L. and Peters,  S. and Pinoy,  N. and Rolnick,  D. and Skinner,  G. L. and Strickson,  O. T. and Svenning,  A. and Teagle,  S. and Høye,  T. T.},
  year = {2024},
  month = may 
}

@article{ChenECE2019,
  title = {Wildlife surveillance using deep learning methods},
  volume = {9},
  ISSN = {2045-7758},
  DOI = {10.1002/ece3.5410},
  number = {17},
  journal = {Ecology and Evolution},
  publisher = {Wiley},
  author = {Chen,  Ruilong and Little,  Ruth and Mihaylova,  Lyudmila and Delahay,  Richard and Cox,  Ruth},
  year = {2019},
  month = aug,
  pages = {9453–9466}
}

@article{MiaoSciRep2019,
  title = {Insights and approaches using deep learning to classify wildlife},
  volume = {9},
  ISSN = {2045-2322},
  DOI = {10.1038/s41598-019-44565-w},
  number = {1},
  journal = {Scientific Reports},
  publisher = {Springer Science and Business Media LLC},
  author = {Miao,  Zhongqi and Gaynor,  Kaitlyn M. and Wang,  Jiayun and Liu,  Ziwei and Muellerklein,  Oliver and Norouzzadeh,  Mohammad Sadegh and McInturff,  Alex and Bowie,  Rauri C. K. and Nathan,  Ran and Yu,  Stella X. and Getz,  Wayne M.},
  year = {2019},
  month = may 
}

@InProceedings{RedmonCVPRYOLO2016,
author = {Redmon, Joseph and Divvala, Santosh and Girshick, Ross and Farhadi, Ali},
title = {You Only Look Once: Unified, Real-Time Object Detection},
booktitle = {Proceedings of the IEEE Conference on Computer Vision and Pattern Recognition (CVPR)},
month = {June},
year = {2016}
}

@misc{BorowiecCDLEcoEvoDL2021,
  title = {Deep learning as a tool for ecology and evolution},
  DOI = {10.32942/osf.io/nt3as},
  publisher = {California Digital Library (CDL)},
  author = {Borowiec,  Marek and Frandsen,  Paul and Dikow,  Rebecca and McKeeken,  Alexander and Valentini,  Gabriele and White,  Alexander},
  year = {2021},
  month = jun 
}

@article{HyePNASEntomologyDL2021,
  title = {Deep learning and computer vision will transform entomology},
  volume = {118},
  ISSN = {1091-6490},
  DOI = {10.1073/pnas.2002545117},
  number = {2},
  journal = {Proceedings of the National Academy of Sciences},
  publisher = {Proceedings of the National Academy of Sciences},
  author = {Høye,  Toke T. and \"{A}rje,  Johanna and Bjerge,  Kim and Hansen,  Oskar L. P. and Iosifidis,  Alexandros and Leese,  Florian and Mann,  Hjalte M. R. and Meissner,  Kristian and Melvad,  Claus and Raitoharju,  Jenni},
  year = {2021},
  month = jan 
}

@software{JocherUltralytics2023,
author = {Jocher, Glenn and Qiu, Jing and Chaurasia, Ayush},
license = {AGPL-3.0},
month = jan,
title = {{Ultralytics YOLO}},
url = {https://github.com/ultralytics/ultralytics},
version = {8.0.0},
year = {2023}
}

@article{BjergeRSECRealTimeInsect2021,
  title = {Real‐time insect tracking and monitoring with computer vision and deep learning},
  volume = {8},
  ISSN = {2056-3485},
  DOI = {10.1002/rse2.245},
  number = {3},
  journal = {Remote Sensing in Ecology and Conservation},
  publisher = {Wiley},
  author = {Bjerge,  Kim and Mann,  Hjalte M. R. and Høye,  Toke Thomas},
  editor = {Sankey,  Temuulen and Ahumada,  Jorge},
  year = {2021},
  month = nov,
  pages = {315–327}
}

@article{KlausBAEImpvBeeMoni2024,
  title = {Improving wild bee monitoring,  sampling methods,  and conservation},
  volume = {75},
  ISSN = {1439-1791},
  DOI = {10.1016/j.baae.2024.01.003},
  journal = {Basic and Applied Ecology},
  publisher = {Elsevier BV},
  author = {Klaus,  Felix and Ayasse,  Manfred and Classen,  Alice and Dauber,  Jens and Diek\"{o}tter,  Tim and Everaars,  Jeroen and Fornoff,  Felix and Greil,  Henri and Hendriksma,  Harmen P. and J\"{u}tte,  Tobias and Klein,  Alexandra Maria and Krahner,  André and Leonhardt,  Sara D. and L\"{u}ken,  Dorothee J. and Paxton,  Robert J. and Schmid-Egger,  Christian and Steffan-Dewenter,  Ingolf and Thiele,  Jan and Tscharntke,  Teja and Erler,  Silvio and Pistorius,  Jens},
  year = {2024},
  month = mar,
  pages = {2–11}
}

@inproceedings{xu2019MissLabelsCVPR,
 author = {Xu, Mengmeng and Bai, Yancheng and Ghanem, Bernard},
 booktitle = {The IEEE Conference on Computer Vision and Pattern Recognition (CVPR) Workshops},
 title = {Missing Labels in Object Detection},
 year = {2019}
}

@misc{LabelStudio,
  title={{Label Studio}: Data labeling software},
  url={https://github.com/HumanSignal/label-studio},
  note={Open source software available from https://github.com/HumanSignal/label-studio},
  author={
    Maxim Tkachenko and
    Mikhail Malyuk and
    Andrey Holmanyuk and
    Nikolai Liubimov},
  year={2020-2025},
}

@article{ZattaraBeeDeclineOneEarth2021,
  title = {Worldwide occurrence records suggest a global decline in bee species richness},
  volume = {4},
  ISSN = {2590-3322},
  DOI = {10.1016/j.oneear.2020.12.005},
  number = {1},
  journal = {One Earth},
  publisher = {Elsevier BV},
  author = {Zattara,  Eduardo E. and Aizen,  Marcelo A.},
  year = {2021},
  month = jan,
  pages = {114–123}
}

@article{WagnerBioDivSurvBSE2022,
  title = {Business,  biodiversity and ecosystem services: Evidence from large‐scale survey data},
  volume = {32},
  ISSN = {1099-0836},
  DOI = {10.1002/bse.3141},
  number = {5},
  journal = {Business Strategy and the Environment},
  publisher = {Wiley},
  author = {Wagner,  Marcus},
  year = {2022},
  month = jun,
  pages = {2583–2599}
}

@misc{EUPoMSProposal2021,
  author       = {{European Commission, Joint Research Centre}},
  title        = {A proposal for an EU Pollinator Monitoring Scheme (EU-PoMS)},
  year         = {2021},
  url          = {https://joint-research-centre.ec.europa.eu},
  note         = {Accessed: 2025-04-05}
}

@article{StaabTrapNestOverviewMEE2018,
  title = {Trap nests for bees and wasps to analyse trophic interactions in changing environments—A systematic overview and user guide},
  volume = {9},
  ISSN = {2041-210X},
  DOI = {10.1111/2041-210x.13070},
  number = {11},
  journal = {Methods in Ecology and Evolution},
  publisher = {Wiley},
  author = {Staab,  Michael and Pufal,  Gesine and Tscharntke,  Teja and Klein,  Alexandra‐Maria},
  editor = {Iossa,  Graziella},
  year = {2018},
  month = aug,
  pages = {2226–2239}
}

@article{TscharntkeBioIndiTrapNestJAE1998,
  title = {Bioindication using trap‐nesting bees and wasps and their natural enemies: community structure and interactions},
  volume = {35},
  ISSN = {1365-2664},
  DOI = {10.1046/j.1365-2664.1998.355343.x},
  number = {5},
  journal = {Journal of Applied Ecology},
  publisher = {Wiley},
  author = {Tscharntke,  Teja and Gathmann,  Achim and Steffan‐Dewenter,  Ingolf},
  year = {1998},
  month = oct,
  pages = {708–719}
}

@article{FornoffDNABarcodingTrapNestICD2023,
  title = {DNA barcoding resolves quantitative multi‐trophic interaction networks and reveals pest species in trap nests},
  volume = {16},
  ISSN = {1752-4598},
  DOI = {10.1111/icad.12664},
  number = {5},
  journal = {Insect Conservation and Diversity},
  publisher = {Wiley},
  author = {Fornoff,  Felix and Halla,  Wenzel and Geiger,  Sarah and Klein,  Alexandra‐Maria and Sann,  Manuela},
  year = {2023},
  month = jul,
  pages = {725–731}
}

@article{LindermannCitizenSciMoniBeeCSTP2024,
  title = {Citizen Science-Based Monitoring of Cavity-Nesting Wild Bees and Wasps–Benefits for Volunteers, Insects, and Ecological Science},
  volume = {9},
  ISSN = {2057-4991},
  DOI = {10.5334/cstp.632},
  number = {1},
  journal = {Citizen Science: Theory and Practice},
  publisher = {Ubiquity Press,  Ltd.},
  author = {Lindermann,  Lara and Grabener,  Swantje and Hellwig,  Niels and Stahl,  Johanna and Dieker,  Petra},
  year = {2024},
  month = sep,
  pages = {22}
}

@article{DuerrbaumMetabarcodingTrapNestME2023,
  title = {Metabarcoding of trap nests reveals differential impact of urbanization on cavity‐nesting bee and wasp communities},
  volume = {32},
  ISSN = {1365-294X},
  DOI = {10.1111/mec.16818},
  number = {23},
  journal = {Molecular Ecology},
  publisher = {Wiley},
  author = {D\"{u}rrbaum,  Ellen and Fornoff,  Felix and Scherber,  Christoph and Vesterinen,  Eero J. and Eitzinger,  Bernhard},
  year = {2023},
  month = jan,
  pages = {6449–6460}
}

@book{lindermannBienenNisthilfen2023,
  address = {DE},
  title = {Wildbienen und Wespen in Nisthilfen bestimmen},
  ISBN = {9783865762627},
  DOI = {10.3220/MX1685523077000},
  publisher = {Johann Heinrich von Th\"{u}nen Institut},
  author = {Lindermann,  Lara and Grabener,  Swantje and Fornoff,  Felix and Hopfenm\"{u}ller,  Sebastian and Schiele,  Susanne and Stahl,  Johanna and Dieker,  Petra},
  year = {2023}
}

@article{Fornoff2025pollenMNU,
  author    = {Yelva Larsen and Maurice Kalweit and Felix Fornoff and Denis Messig},
  title     = {Pollen oder Grashüpfer? : Digitale Bestimmung von Wildbienen und anderen Insekten im Unterricht},
  journal   = {MNU Journal},
  volume    = {78},
  number    = {1},
  pages     = {13--17},
  year      = {2025},
  publisher = {Verlag Klaus Seeberger},
  address   = {Neuss},
  issn      = {0025-5866}
}

@misc{KalweitCitizenSciWildbienenBamberg2024,
  title = {Citizen-Science in der Lehre durch eine digitale Wildbienenbestimmung},
  DOI = {10.20378/irb-94524},
  publisher = {Universitatsbibliothek Bamberg},
  author = {Kalweit,  Maurice and Raab,  Patricia and Larsen,  Yelva and Fornoff,  Felix and Messig,  Denis},
  year = {2024}
}

@article{AlkassabEcotoxiComparBee2020,
  title = {Comparing response of buff-tailed bumblebees and red mason bees to application of a thiacloprid-prochloraz mixture under semi-field conditions},
  volume = {29},
  ISSN = {1573-3017},
  DOI = {10.1007/s10646-020-02223-2},
  number = {7},
  journal = {Ecotoxicology},
  publisher = {Springer Science and Business Media LLC},
  author = {Alkassab,  Abdulrahim T. and Kunz,  Nadine and Bischoff,  Gabriela and Pistorius,  Jens},
  year = {2020},
  month = may,
  pages = {846–855}
}

@article{CalvusEIInfieldMoniGroundNest2025,
  title = {In-field monitoring of ground-nesting insect aggregations using a scaleable multi-camera system},
  volume = {86},
  ISSN = {1574-9541},
  DOI = {10.1016/j.ecoinf.2025.103004},
  journal = {Ecological Informatics},
  publisher = {Elsevier BV},
  author = {Calvus,  Daniela and Wueppenhorst,  Karoline and Schl\"{o}sser,  Ralf and Klaus,  Felix and Schwanecke,  Ulrich and Greil,  Henri},
  year = {2025},
  month = may,
  pages = {103004}
}

@book{spellerberg2005monitoring,
  title={Monitoring Ecological Change},
  author={Spellerberg, I.F.},
  isbn={9780521820288},
  lccn={2006271065},
  year={2005},
  publisher={Cambridge University Press}
}

@inproceedings{LiuVISAPP2026,
  title = {Order-Level Arthropod Detection Using Deep Learning: Addressing Scale Variability through Synthetic Data},
  url = {http://dx.doi.org/10.5220/0014220300004084},
  DOI = {10.5220/0014220300004084},
  booktitle = {Proceedings of the 21st International Conference on Computer Vision Theory and Applications},
  publisher = {SCITEPRESS - Science and Technology Publications},
  author = {Liu,  Chenchang and Ionova,  Svetlana and M\"{a}der,  Patrick and Seeland,  Marco},
  year = {2026},
  pages = {210–217}
}

@inproceedings{lin2017focal,
  title={Focal loss for dense object detection},
  author={Lin, Tsung-Yi and Goyal, Priya and Girshick, Ross and He, Kaiming and Doll{\'a}r, Piotr},
  booktitle={Proceedings of the IEEE international conference on computer vision},
  pages={2980--2988},
  year={2017}
}

@misc{chatgpt2026,
  author       = {OpenAI},
  title        = {GPT},
  year         = {2026},
  howpublished = {\url{https://chat.openai.com}},
  note         = {Accessed: 2026-05-26}
}
\end{document}